\documentclass[11pt,conference]{IEEEtran}
\usepackage{cite}
\usepackage{amsmath,amssymb,amsfonts}
\usepackage{algorithmic}
\usepackage{graphicx}
\usepackage{booktabs} 
\usepackage{longtable}
\usepackage{textcomp}
\usepackage{xcolor}
\usepackage[utf8]{inputenc}
\usepackage[english]{babel}
\usepackage{hyperref}
\usepackage{url}

\begin{document}

\title{Enhancing Road Safety: Real-Time Detection of Driver Distraction through Convolutional Neural Networks}

\author{
\IEEEauthorblockN{Amaan Aijaz Sheikh}
\IEEEauthorblockA{\textit{NetId: asheikh4}\\
asheikh4@illinois.edu}
\and
\IEEEauthorblockN{Imaad Zaffar Khan}
\IEEEauthorblockA{\textit{NetId: izkhan2}\\
izkhan2@illinois.edu}
}

\maketitle

\begin{abstract}
As we navigate our daily commutes, the threat posed by a distracted driver is at a large, resulting in a troubling rise in traffic accidents. Addressing this safety concern, our project harnesses the analytical power of Convolutional Neural Networks (CNNs), with a particular emphasis on the well-established models VGG16 and VGG19. These models are acclaimed for their precision in image recognition and are meticulously tested for their ability to detect nuances in driver behavior under varying environmental conditions. Through a comparative analysis against an array of CNN architectures, this study seeks to identify the most efficient model for real-time detection of driver distractions. The ultimate aim is to incorporate the findings into vehicle safety systems, significantly boosting their capability to prevent accidents triggered by inattention. This research not only enhances our understanding of automotive safety technologies but also marks a pivotal step towards creating vehicles that are intuitively aligned with driver behaviors, ensuring safer roads for all.

\end{abstract}

\section{Introduction}
\subsection{Background}
Distracted driving has emerged as a significant safety hazard on roads worldwide, contributing to a great number of traffic accidents each year. With the advent of advanced computational technologies, there is a promising potential to mitigate these risks through the development of real-time detection systems. Convolutional Neural Networks (CNNs), renowned for their effectiveness in image recognition tasks, provide a foundational technology for analyzing complex visual behaviors associated with driving.

\subsection{Problem Statement}
Despite ongoing efforts to enhance vehicular safety, existing systems find it difficult to handle the subtleties of driver distraction in various driving conditions and environments. Current technologies often rely on simplistic alert mechanisms or invasive monitoring techniques that may not accurately detect or predict momentary lapses in driver attention. This project seeks to address these limitations by employing a sophisticated CNN-based approach to detect and quantify driver distractions more effectively and unobtrusively.

\subsection{Objectives}

The central aim of our project is to conduct a comprehensive analysis and testing of various CNN architectures for real-time detection of driver distraction. This includes:

\begin{itemize}
    \item Evaluating simple CNN architectures to establish baseline performance metrics.
    \item Assessing both batchwise and non-batchwise fine-tuning of VGG16 and VGG19 models to determine their effectiveness in different processing environments.
    \item Comparing the impacts of shallow versus deep configurations of VGG16 and VGG19 networks on the accuracy of distraction detection.
    \item Developing and testing a custom CNN architecture with a transformer to tailor the detection system to the specific dynamics of driver behavior under varied conditions.
\end{itemize}

By exploring these methodologies, the project aims to provide a thorough understanding of how different CNN models can enhance the detection of driver distraction and thereby contribute significantly to road safety. 

Following this section, a comprehensive related work section will contextualize our findings within the broader field of automotive safety technologies, comparing our innovative methods with existing approaches and setting a foundation for the subsequent implementation and evaluation phases.

\section{Related Work}
In this section, we delve into significant prior studies that explore the use of convolutional neural networks in detecting driver distraction. Each of these studies contributes to the foundation upon which our research is built, and comparing their methodologies and findings with ours offers valuable insights into the evolution of this technology.

\subsection{Automatic Driver Distraction Detection Using Deep Convolutional Neural Networks}
Md. Uzzol Hossain and colleagues (2022) \cite{hossain2021} employed deep convolutional neural networks to identify driver distraction automatically. Their work utilized a series of complex models to process visual data from in-vehicle cameras. This study's strength lies in its extensive dataset and the depth of its neural network analysis, which provides a solid benchmark for our project. However, our approach extends this by incorporating newer models like VGG19 and exploring batchwise versus non-batchwise processing to enhance real-time application capabilities.

\subsection{Detection of Distracted Driver Using Convolution Neural Network}
The 2022 study by Narayana Darapaneni \cite{darapaneni2021} outlines a CNN-based framework for detecting driver distraction that focuses on processing constraints in real-time  Their methodology is aligned with our work, particularly in their use of a streamlined model for efficient computation. The insights from this study guide our exploration of computational efficiency in model training and real-time detection, providing a comparative perspective that enriches our approach.

\subsection{Detecting Distraction of Drivers Using Convolutional Neural Network} 
In this study, Sarfaraz Masood and his team (2018) \cite{masood2018} tackled the challenge of detecting driver distraction using a tailored CNN. This early work is crucial as it sets a methodological precedent for subsequent studies. Their findings, particularly regarding network configuration and the impact of non-standard data pre-processing techniques, have influenced our decision to experiment with various pre-processing methods and architectural tweaks to optimize performance.

\vspace{0.10cm}

Each of these studies contributes uniquely to the body of knowledge in detecting driver distraction via CNNs. By integrating their empirical findings with our methodological innovations, this literature review not only frames our research within the current scientific dialogue but also sets the stage for our contributions to advance the field further.

\section{Dataset}

The dataset used in our project is the "State Farm Distracted Driver Detection,"\cite{dataset} available through Kaggle. This dataset is pivotal for training and testing our convolutional neural network models to accurately identify different types of driver distractions and is considered a standard when dealing with data for driver distraction training. It consists of images categorized into ten classes, each representing a specific form of distraction. Below we can see the description/category name of each of the classes and the count of their images in the dataset:

\begin{table}[h]
\centering
\caption{Distribution of Classes in the State Farm Distracted Driver Detection Dataset}
\label{tab:dataset_classes}
\begin{tabular}{|c|l|c|}
\hline
\textbf{Label} & \textbf{Description}                 & \textbf{Count of Images} \\ \hline
c0             & Safe driving                         & 2489           \\ \hline
c1             & Texting - right                      & 2267           \\ \hline
c2             & Talking on the phone - right         & 2317           \\ \hline
c3             & Texting - left                       & 2346           \\ \hline
c4             & Talking on the phone - left          & 2326           \\ \hline
c5             & Operating the radio                  & 2312           \\ \hline
c6             & Drinking                             & 2325           \\ \hline
c7             & Reaching behind                      & 2002           \\ \hline
c8             & Hair and makeup                      & 1911           \\ \hline
c9             & Talking to passenger                 & 2129           \\ \hline
\end{tabular}
\end{table}

As can be seen from above, the summation of the count of images leads to the total dataset having 22424 images in total.
The distribution of images across these categories is relatively balanced, with the counts ranging from 1911 for 'hair and makeup' to 2489 for 'safe driving,' ensuring that our model trains on a diverse set of data.

\subsection{Initial Data Analysis}

An initial analysis of the dataset was conducted to understand the properties of the images it contains. The color distribution of pixel intensities across the RGB channels was examined, with the following observations:

\begin{figure}
    \centering
    \includegraphics[width=0.5\textwidth]{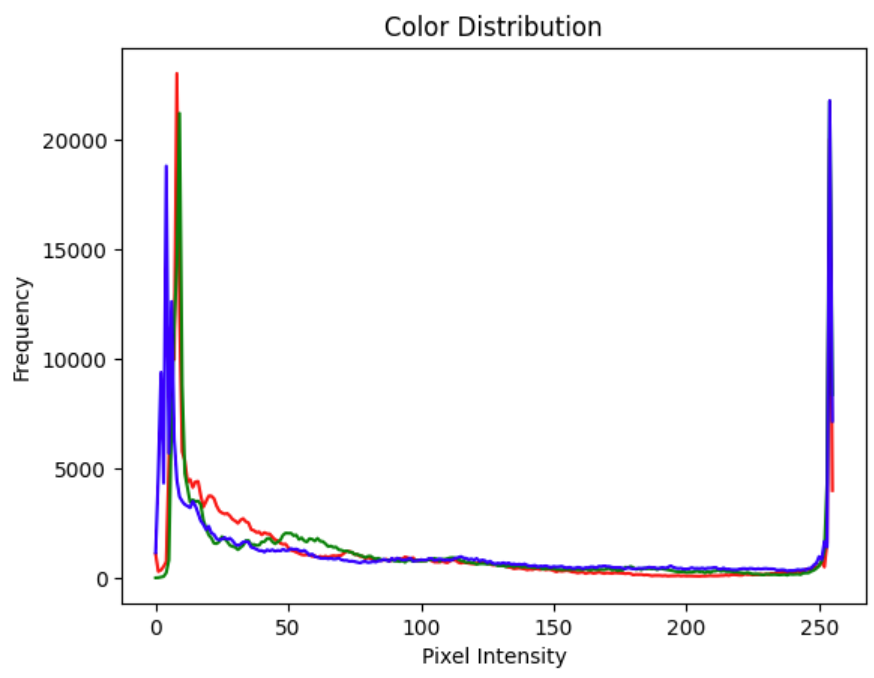}
    \caption{Pixel Intensity Distribution across RGB channels}
\end{figure}

\begin{itemize}
    \item \textbf{Peaks at the Edges:} The figure shows prominent spikes at the minimum and maximum intensity values (0 and 255), indicating significant occurrences of very dark and very bright pixels. This suggests potential issues with overexposure or underexposure in various images.
    \item \textbf{Middle Lows:} There is a noticeable lack of pixel intensity values in the mid-range, which can imply a loss of detail in mid-tones, possibly affecting the richness of visual information.
    \item \textbf{High Contrast:} The pronounced peaks at the edges combined with lower frequencies in the mid-range highlight a high-contrast nature in the images, which could enhance the differentiation of important features but may also lead to challenges in distinguishing subtler details.
    \item \textbf{Balanced Colors:} The alignment of red, green, and blue channels in the distribution pattern suggests that the images are well-balanced in terms of color, not overly dominated by any single color channel.
\end{itemize}

This preliminary analysis is critical as it informs the necessary pre-processing steps we must undertake to standardize the images for optimal CNN performance, such as adjusting brightness levels and enhancing contrast to ensure that important features are well-represented.

\section{Methodology}
\subsection{System Design and Code Structuring}
Our project makes use of a system we designed to maximize flexibility and efficiency in model experimentation. Central to our approach is a streamlined code structure that allows for the seamless integration of various CNN architectures with minimal changes required between experiments.

The base of our code remains consistent across all models, taking care of data handling, pre-processing, and augmentation, as well as the configuration of the training environment. This design simplifies the process of switching between different neural network models, ensuring that each model is evaluated under the same conditions to maintain consistent comparison standards in terms of hyperparameter and the overall environment.

\textbf{Consistent Code Base:} The following components are standardized in our system:
\begin{itemize}
    \item \textbf{Data Pre-processing and Augmentation:} We utilize an \texttt{ImageDataGenerator} for real-time data augmentation, enhancing model robustness by simulating various real-world distortions.
    \item \textbf{Model Training and Evaluation Setup:} Model checkpointing and early stopping are implemented to optimize training phases and prevent overfitting, ensuring that each model's performance is maximized.
    \item \textbf{Performance Visualization:} After training, we plot the training and validation/test accuracy and loss to assess model performance over epochs.
\end{itemize}

\textbf{Variable Components:}
Model initialization functions are the primary variable in our system. By designing the code to alter only the model initialization segment, we can easily deploy different CNN architectures, such as variants of VGG16, VGG19 and pretty much all the models we would want to try, without affecting other parts of the code base.

Based on thye above discussion, below we can see an highlevel overview of what our system looks like:
\begin{figure}[h]
    \centering
    \includegraphics[width=0.5\textwidth]{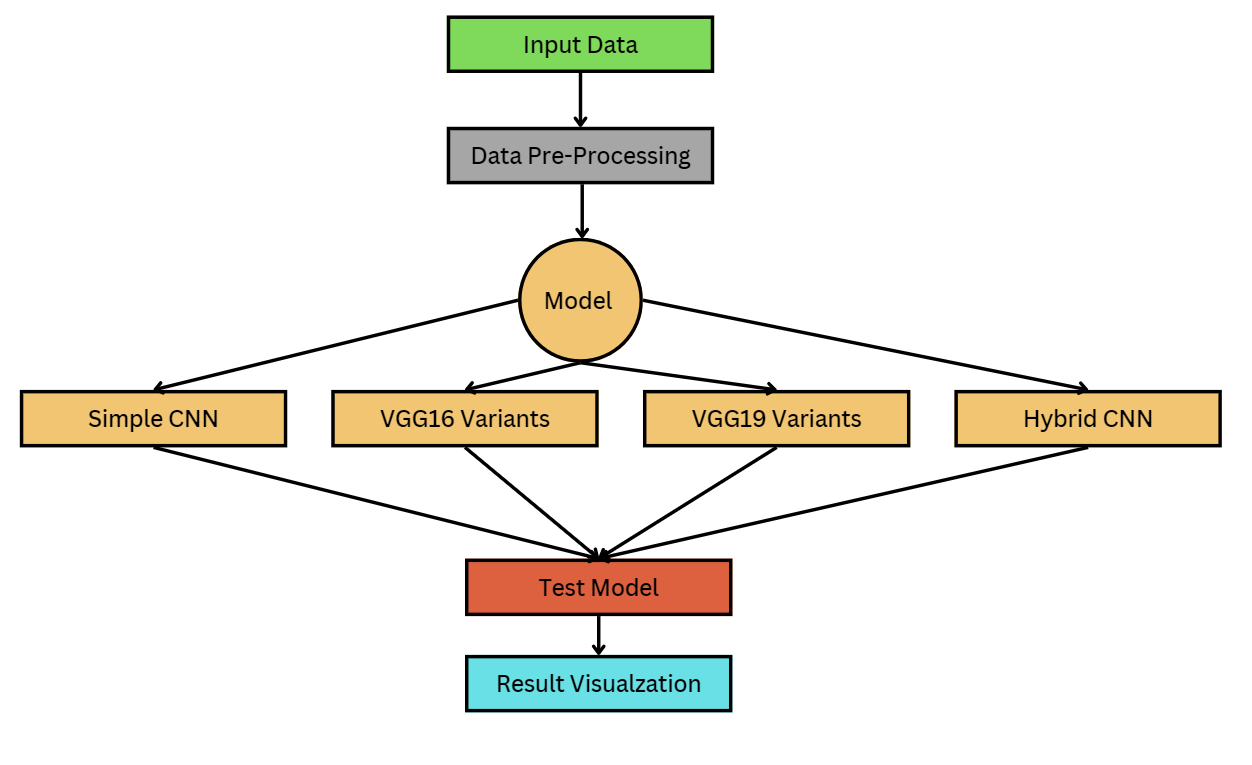}
    \caption{System Architecture}
\end{figure}

As can be seen above, the data-preprocessed is ideally going to remain constant across the board, with the changes only being made to the model functions. 
We talk about two other key features in the subsections below:

\subsubsection{Model Experimentation}
Each model is instantiated using a specific initialization function, which defines the architecture and compilation settings. This approach allows for straightforward comparison across models, focusing on:
\begin{itemize}
    \item \textbf{Architecture Efficiency:} By comparing the different architectures, we assess the trade-offs between computational demand and predictive performance.
    \item \textbf{Hyperparameter Consistency:} Hyperparameters such as learning rates and batch sizes are kept consistent, except where adjustments are necessitated by early stopping criteria based on validation loss.
\end{itemize}

\subsubsection{Implementation of Early Stopping}
Early stopping is a critical component of our training strategy, used to halt the training process if the validation loss does not improve for a predefined number of epochs. This mechanism not only saves computational resources but also aids in preventing the overfitting of models to the training data.

\subsection{Data Preprocessing and Augmentation}
The effectiveness of our Convolutional Neural Network models is heavily dependent on the quality and variety of the data they are trained on. To ensure our models  capable of generalizing well to real-world scenarios, we implement comprehensive data pre-processing and augmentation techniques. These techniques are designed to simulate realistic variations and distortions that might occur in real driving conditions.

\textbf{Pre-processing Functions:}
Our pre-processing pipeline includes functions to dynamically enhance the brightness and contrast of the input images. This is achieved through the following steps:
\begin{itemize}
    \item \textbf{Enhance Brightness:} The brightness of images is adjusted by converting the images to the HSV color space and manipulating the value channel. This helps in simulating different lighting conditions that a driver might experience.
    \item \textbf{Change Contrast:} The contrast of images is modified using a scaling factor to adjust the intensity of each pixel. This ensures that the model will be able to identify relevant features under different possible visual contrasts.
\end{itemize}

These pre-processing functions are integrated into our data generators as callable transformations, which apply a random combination of brightness and contrast adjustments to each image during training and validation. This randomness helps in enhancing the model's ability to generalize over diverse visual representations.

\textbf{Augmentation Techniques:}
To further augment our dataset, we employ several image transformation techniques, which include:
\begin{itemize}
    \item \textbf{Random Rotations} and \textbf{Horizontal Flips} to simulate different angles and orientations of the driver relative to the camera.
    \item \textbf{Random Affine Transformations} provide slight translations, shears, and scaling to replicate the effect of camera shifts and zooms.
    \item \textbf{Randomized Brightness and Contrast Adjustments} are applied to ensure diversity in lighting and exposure.
\end{itemize}

We utilize two distinct setups for our data handling:
\begin{itemize}
    \item \textbf{Keras ImageDataGenerator:} For experiments utilizing TensorFlow and Keras, we configure an \texttt{ImageDataGenerator} that applies our pre-processing functions in real-time during model training, reducing memory overhead and introducing realistic variations.
    \item \textbf{Torchvision Transforms:} For PyTorch-based implementations, we use a composition of transforms that include custom classes for brightness and contrast adjustments alongside standard transformations provided by \texttt{torchvision}.
\end{itemize}
The idea behind using two different libraries to essentially get the same requirement output was done in order to get comfortable with both the libraries and solely for educational purposes. In reality, both the above mentioned data handlers would in essence provide an output similar to each other.

These pre-processing and augmentation strategies are crucial for training deep learning models that are effective and reliable in diverse and unpredictable real-world driving conditions. They not only enhance the data's variability but also improve the model's robustness to overfitting, thereby increasing its practical applicability in real-time systems.

\subsection{Model Initialization and Training}

Until now, we have looked at the very initial phases of our system, now we delve into the depths of understanding the various models that we compared, have an understanding about them, what is the architecture that they had post which in the results section we will have an extensive analysis on the performance of each of the models followed by a comparison of the testing we did.

The following sections will feature the various models that we tried and tested, and the results we were able to produce.

\section{Model Architectures}
In this section, we have a look at the various CNN architectures developed and evaluated as part of our studys. The diversity in model architectures is intended to explore a range of complexities and computational efficiencies, thereby identifying the most suitable model for real-time application in diverse driving environments.

\subsection{SimpleCNN}
The SimpleCNN model represents our baseline architecture. It is designed to be lightweight yet effective, suitable for environments where computational resources are limited.

\subsubsection{Architecture}
The SimpleCNN comprises three convolutional layers, each followed by a ReLU activation and a max pooling operation. The overall architecture of the CNN can be seen below:
\begin{figure}[h]
    \centering
    \includegraphics[width=0.45\textwidth]{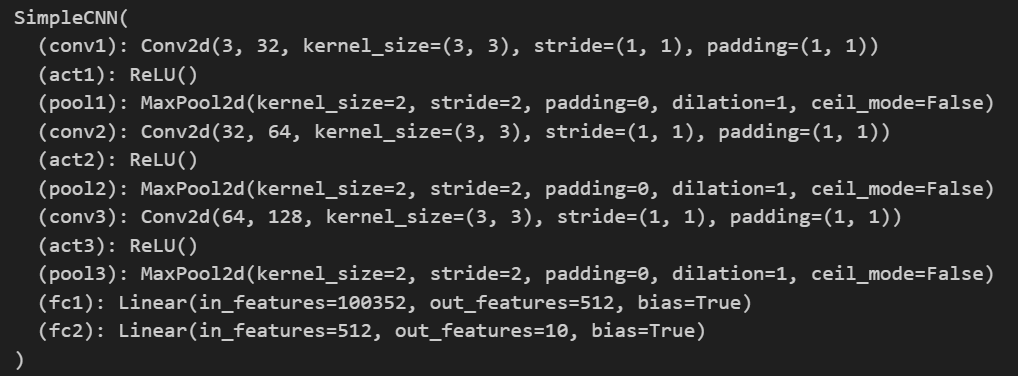}
    \caption{System Architecture}
\end{figure}

Here is the breakdown based on the above figure:
\begin{itemize}
    \item \textbf{First Layer:} The initial layer uses 32 filters with a kernel size of 3x3 and has a padding to maintain the size of the output feature map. A max pooling operation with a stride of 2 reduces the spatial dimensions by half.
    \item \textbf{Second Layer:} This layer increases the filter count to 64, enhancing the network's ability to capture more complex features from the input images. It follows the same pattern of convolution, activation, and pooling.
    \item \textbf{Third Layer:} The complexity increases further with 128 filters. This layer is crucial for extracting fine-grained details that are essential for accurate classification.
\end{itemize}

Following the convolutional base, the network transitions to fully connected layers:
\begin{itemize}
    \item \textbf{Flatten Layer:} The output from the last pooling layer is flattened into a vector to serve as input for the dense layers.
    \item \textbf{Fully Connected Layers:} A sequence of two dense layers processes the flattened features, with the first layer consisting of 512 units followed by the final output layer that maps to the number of classes (10 in this case).
\end{itemize}

\subsubsection{Rationale}
SimpleCNN is designed as a straightforward and computationally efficient model that can be easily deployed in real-time systems with limited hardware capabilities. Its simplicity allows for rapid training and inference, making it an ideal candidate for initial experiments and as a benchmark for more complex architectures. By using a moderate number of layers and parameters, it balances the trade-off between performance and computational demand, making it suitable for scenarios where real-time processing speed is critical.

\subsection{VGG16 Architectures}
The VGG16 architecture, known for its effectiveness in large-scale image recognition tasks, was chosen for its depth and robust feature extraction capabilities, making it highly suitable for complex tasks like this one. We experimented with three distinct implementations of this architecture: a deep model, a shallow model, and fine-tuning adaptations, to explore how different configurations impact performance.

\subsubsection{Deep VGG16 Model}

The deep configuration uses the VGG16 model as a base, extended with additional dense layers to increase the model's capacity for learning from our specific dataset.
\begin{itemize}
    \item The model begins with the VGG16 base, pre-trained on ImageNet, excluding the top layer to tailor the outputs to our ten classes.
    \item It includes a Flatten layer to transform the feature maps into a 1D array.
    \item Followed by a dense layer with 500 units and ReLU activation for non-linear transformation.
    \item A dropout of 0.5 is included to prevent overfitting.
    \item The output layer is a dense layer with softmax activation suited for multi-class classification.
\end{itemize}

\textbf{Rationale:}
This model is designed to utilize the deep feature extraction capabilities of VGG16, enhancing it with additional trainable parameters to better adapt to the intricate nature of driver behavior recognition.

\subsubsection{Shallow VGG16 Model}

The shallow model modifies the deep VGG16 by reducing the complexity of the layers added to the base, aiming to maintain efficiency while still capturing essential features.
\begin{itemize}
    \item Uses the same VGG16 base but sets it as non-trainable to focus learning in the newly added layers.
    \item The model includes a single dense layer with 256 units, which is fewer than in the deep model, reducing the learning capacity but also the risk of overfitting.
    \item Includes dropout and softmax activation as in the deep model.
\end{itemize}

\textbf{Rationale:}
The shallow model is particularly valuable for scenarios where computational resources are limited or when quicker inference is needed without a substantial drop in accuracy. It tests whether a less complex model can efficiently handle the task.

\subsubsection{Fine-Tuned VGG16 Model}

Fine-tuning allows us to tailor pre-trained networks to our specific task more finely. This setup involved selectively retraining the deeper layers of the VGG16 network.
\begin{itemize}
    \item The base model’s top layers are fine-tuned while earlier layers remain frozen, balancing the learning of high-level features without altering the robust initial feature detection.
    \item Includes GlobalAveragePooling to reduce dimensionality and manage model complexity.
    \item The model is compiled with an SGD optimizer, suitable for fine-tuning due to its finer control over learning rates.
\end{itemize}

\textbf{Rationale:}
Fine-tuning was tested in both batch-wise and non-batch-wise configurations to assess the impact of training dynamics on performance. Batch-wise fine-tuning restricts adjustments to the last few layers, conserving most of the pre-trained features, while non-batch-wise fine-tuning allows the entire network to adjust to new data.

\textbf{Purpose of Fine-Tuning:}
This approach was chosen to explore how much of the pre-trained model's knowledge could be preserved while still adapting to the specific challenges of detecting driver distractions, potentially leading to improved accuracy with minimal training time compared to training a model from scratch.

These variations of the VGG16 architecture provide a broad spectrum of insights into how different depths and training strategies can be optimized for the task of driver distraction detection.

\subsection{VGG19 Architectures}
Building on the insights from VGG16, we extended our experiments to include VGG19, a deeper variant of the VGG architecture known for its enhanced performance in image recognition tasks. We implemented three versions of this architecture, similar to that of VGG16: a deep model, a shallow model, and fine-tuning adaptations to explore their effectiveness for driver distraction detection.

\subsubsection{Deep VGG19 Model}

The deep VGG19 model incorporates the VGG19 base, known for its depth and complexity, to capture more detailed feature representations.
\begin{itemize}
    \item Starts with the VGG19 base model pre-trained on ImageNet, with the top layers removed to adapt to our specific classification needs.
    \item Includes a Flatten layer to convert the feature maps into a vector.
    \item A dense layer with 500 units followed by a ReLU activation introduces non-linearity and capacity to learn complex patterns.
    \item Dropout of 0.5 is incorporated to reduce overfitting.
    \item The model concludes with a softmax activation layer tailored for our 10-class detection task.
\end{itemize}

\textbf{Rationale:}
This model leverages the deeper network structure of VGG19 to process more complex image features, which is expected to improve the accuracy in distinguishing between various types of driver distractions.

\subsubsection{Shallow VGG19 Model}

The shallow model uses the VGG19 architecture but simplifies the layers added after the base to maintain efficiency.
\begin{itemize}
    \item Utilizes the VGG19 base with frozen weights to focus learning in the newly introduced simple layers.
    \item A single dense layer with 256 units is used, reducing the complexity and computational demand compared to the deep model.
    \item Maintains dropout to avoid overfitting and uses softmax for the classification output.
\end{itemize}

\textbf{Rationale:}
The shallow VGG19 model is designed to test the hypothesis that less complexity might still yield high accuracy, particularly beneficial in environments where computational resources or response time are constraints.

\subsubsection{Fine-Tuned VGG19 Model}

Fine-tuning is applied to the VGG19 base to tailor the deeper layers to our dataset while keeping the initial layers fixed.
\begin{itemize}
    \item Fine-tuning starts from the fourth-last layer, with earlier layers left unchanged to preserve their pre-trained feature-detection capabilities.
    \item Incorporates GlobalAveragePooling to reduce the feature dimensions effectively.
    \item Ends with a dense layer for class prediction.
\end{itemize}

\textbf{Rationale:}
Fine-tuning was implemented in both batch-wise and non-batch-wise settings to evaluate how different training strategies affect the model’s ability to adapt to the specific task of detecting driver distractions. Batch-wise fine-tuning is particularly useful for maintaining the integrity of most learned features while updating only the most crucial layers for the task.

\textbf{Purpose of Using VGG19:}
Choosing VGG19 allowed us to assess the impacts of using a deeper network compared to VGG16, providing insights into whether additional depth translates to better performance for this specific application. It also serves as a comparative study to determine the optimal balance between depth and computational efficiency in practical applications.

These VGG19 models expand our understanding of how varying depths and fine-tuning approaches can be optimized for enhancing the detection capabilities in driver distraction tasks.

\subsection{Hybrid CNN-Transformer Architecture}
The culmination of our exploration into neural network architectures involves a sophisticated Hybrid CNN-Transformer model. This model combines the strengths of CNNs and the transformer architecture to create a powerful tool for image-based classification tasks.

\subsubsection{Architecture Overview}
The Hybrid CNN-Transformer architecture is designed to leverage the detailed feature extraction capabilities of CNNs with the global context retention of transformers. The specifics of the architecture are as follows:
\begin{itemize}
    \item \textbf{ResNet Backbone:} The model utilizes a pre-trained ResNet50 as the CNN component, known for its effectiveness in image recognition. The final fully connected layer of the ResNet50 is removed to prepare the feature maps for the transformer.
    \item \textbf{Dimension Reduction:} A linear layer reduces the dimensionality of the ResNet50’s output from 2048 to 512, making it suitable for processing by the transformer.
    \item \textbf{Transformer Encoder:} The transformer component consists of an encoder with multiple layers, each containing multi-headed self-attention mechanisms that help the model understand the global dependencies within the image.
    \item \textbf{Classification Head:} The final output of the transformer encoder is processed through an average pooling layer followed by a fully connected layer to predict the class labels.
\end{itemize}

\subsubsection{Rationale for Hybrid Architecture}
This hybrid model is crafted to address specific challenges in driver distraction detection:
\begin{enumerate}
    \item \textbf{Feature Extraction:} The ResNet50 backbone captures complex spatial hierarchies in the image data, which are crucial for identifying various forms of driver distraction.
    \item \textbf{Contextual Awareness:} The transformer part of the model integrates these features over the entire image, considering both local and global contexts, which is vital for understanding scenarios where multiple distractions might be present.
    \item \textbf{Flexibility and Depth:} Combining CNNs with transformers allows the model to not only recognize but also interpret the significance of various features across different parts of the image, enhancing its predictive accuracy.
\end{enumerate}

\subsubsection{Implementation Details}

This architecture represents a forward-thinking approach in neural network design for image classification, combining proven techniques with innovative structures to tackle the complex problem of detecting driver distractions effectively.

\subsection{Framework Utilization Rationale}

In our project, we opted to implement our preprocessing and augmentation pipelines as well as some models in both Keras (using TensorFlow as the backend) and PyTorch. This dual-implementation approach was chosen deliberately to achieve a comprehensive understanding of the functionalities and advantages of both leading deep learning libraries.

\textbf{Consistency Across Frameworks:}
Although the implementations differ in syntax and library-specific functions, the core pre-processing and augmentation functionalities are maintained consistently across both Keras and PyTorch. This ensures that the output—augmented and pre-processed images—is equivalent, regardless of the framework used. This consistency allows for an unbiased comparison of model performance that is strictly attributable to the model architectures and not influenced by data variations.

\textbf{Educational Value:}
The choice to utilize both frameworks also stems from an educational perspective. By exploring the same tasks with different tools, we gain insights into the strengths and limitations of each framework, a simpel example of which is the progress bar of the epochs in keras being default, while we had to have a tqdm wrapper fro PyTorch. This experience is invaluable for our team’s skill development and provides a broader perspective on solving machine learning problems in practice.

\textbf{Practical Implications:}
From a practical standpoint, understanding how to implement similar tasks in different frameworks enhances our flexibility and adaptability in the field. It prepares our team to work with diverse technologies and adapt to various technical environments in future projects or professional settings.

This dual-framework approach not only enriches our technical proficiency but also ensures that our findings and conclusions are robust, backed by the capability to replicate results across different software environments, thereby reinforcing the reliability and validity of our research outcomes..

\section{Training and Validation Results}
This subsection presents the training and validation results for each of the models discussed: SimpleCNN, VGG16 and subtypes, VGG19 and subtypes, and Hybrid CNN-Transformer. We evaluate each model's performance based on accuracy, loss, and other relevant metrics over the course of the training epochs. This analysis helps in understanding how each model learns and generalizes from the training data to the validation data set.
It is to be noted that environmental parameters like learning rate, epochs, early stopping, patience etc were kept constant across all models.
\subsection{Performance Metrics}
The performance of each model was evaluated using the following metrics calculated during training and validation phases:

\textbf{Average Loss Calculation:}
\[
\text{Average Loss} = \frac{\sum (\text{Loss per Batch} \times \text{Batch Size})}{\text{Total Number of Samples}}
\]

\textbf{Accuracy Calculation:}
\[
\text{Accuracy} = \frac{\text{Number of Correct Predictions}}{\text{Total Number of Samples}}
\]

\subsection{Tabular Summary}
The performance of various models during the training and validation phases is presented in the following table:

\begin{table}[h]
\centering
\caption{Training and Validation Performance of Models}
\label{tab:model_performance1}
\begin{tabular}{|c|c|c|c|c|}
\hline
\textbf{Model} & \textbf{TrainAcc} & \textbf{TrainLoss} & \textbf{ValAcc} & \textbf{ValLoss} \\ \hline
Simple CNN     & 0.9300 & 0.0363 & 0.9500 & 0.5900 \\ \hline
VGG16 Deep     & 0.9952 & 0.0153 & 0.9933 & 0.0300 \\ \hline
VGG16 Shallow  & 0.9612 & 0.0388 & 0.9937 & 0.0600 \\ \hline
VGG16 FT-B & 0.9905 & 0.0438 & 0.9871 & 0.0540 \\ \hline
VGG16 FT-NB & 0.9943 & 0.0246 & 0.9895 & 0.0376 \\ \hline
VGG19 Deep     & 0.9952 & 0.0153 & 0.9949 & 0.0233 \\ \hline
VGG19 Shallow  & 0.9592 & 0.1372 & 0.9920 & 0.0345 \\ \hline
VGG19 FT-B & 0.9921 & 0.0340 & 0.9868 & 0.0475 \\ \hline
VGG19 FT-NB & 0.9943 & 0.0246 & 0.9895 & 0.0376 \\ \hline
Hybrid Model   & 0.9919 & 0.0280 & 0.9918 & 0.0358 \\ \hline

\end{tabular}
\end{table}

This table provides a comprehensive overview of how each model performed in terms of accuracy and loss during the training and validation stages. Graphs and more detailed discussions about each model's performance trends and anomalies follow in subsequent section.
A breif note, unless stated otherwise, the x-axis of the graph denote the epochs while the y-axis denote either of the accuracy or the loss.

\subsection{SimpleCNN}
The simpleCNN design was supposed to act as a baseline, as it can be seen from the below trend in the training and validation loss, the model overall performs as expected, with a consistent decline in the loss values as training progresses.

\begin{figure}[h]
    \centering
    \includegraphics[width=0.4\textwidth]{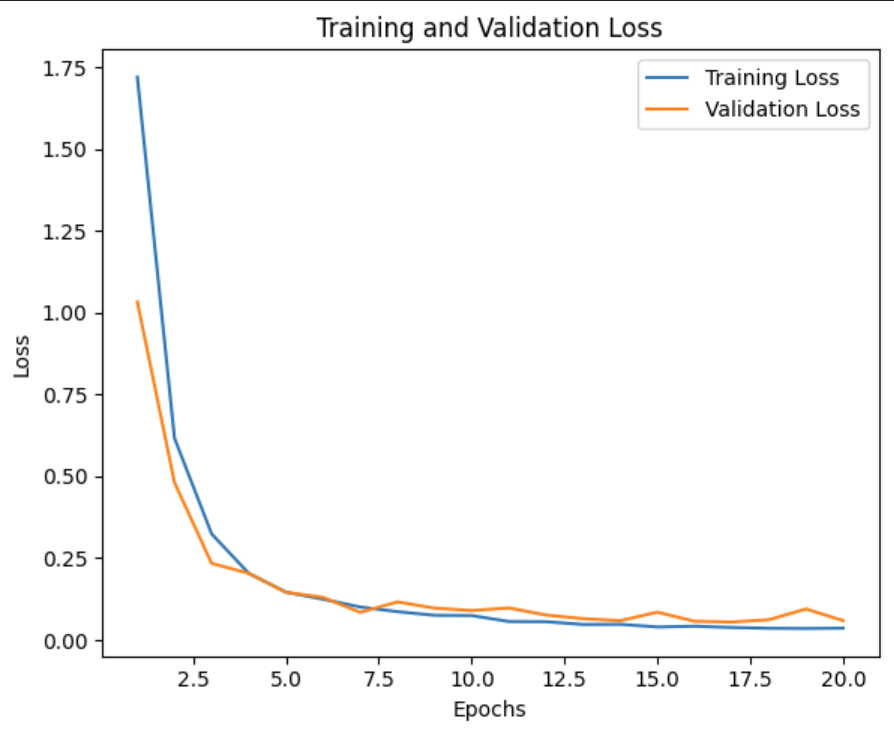}
    \caption{Training and Validation Loss for SimpleCNN}
    \label{fig:simple_cnn_results}
\end{figure}

\subsection{VGG16 - Deep Network}
The VGG16 Deep network as mentioned in the previous sections is quite well proven for image recognition tasks, the architecture we implemented however, is extended with additional dense layers to increase the learning capacity towards our dataset.

Based, on the graphs below for training and validation accuracies as well as losses, it can be seen that the performance of the models was quite steady and consistent with what was expected, however, it is worth mentioning that the training time for the same was quite drastically increased. 
\begin{figure}[h]
    \centering
    \includegraphics[width=0.4\textwidth]{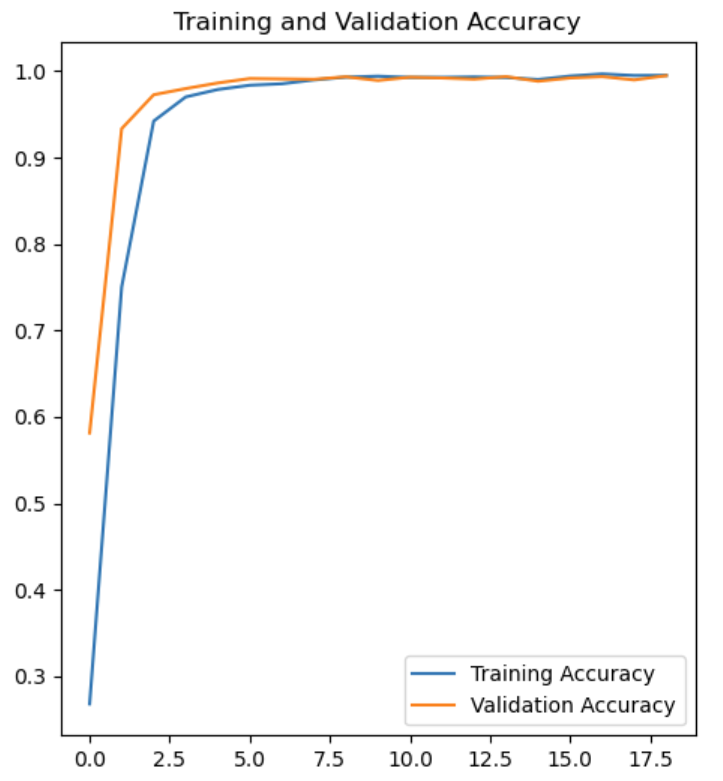}
    \caption{Training and Validation Accuracy VGG16- Deep Network}
    \label{fig:vgg16_results}
\end{figure}
\begin{figure}[h]
    \centering
    \includegraphics[width=0.4\textwidth]{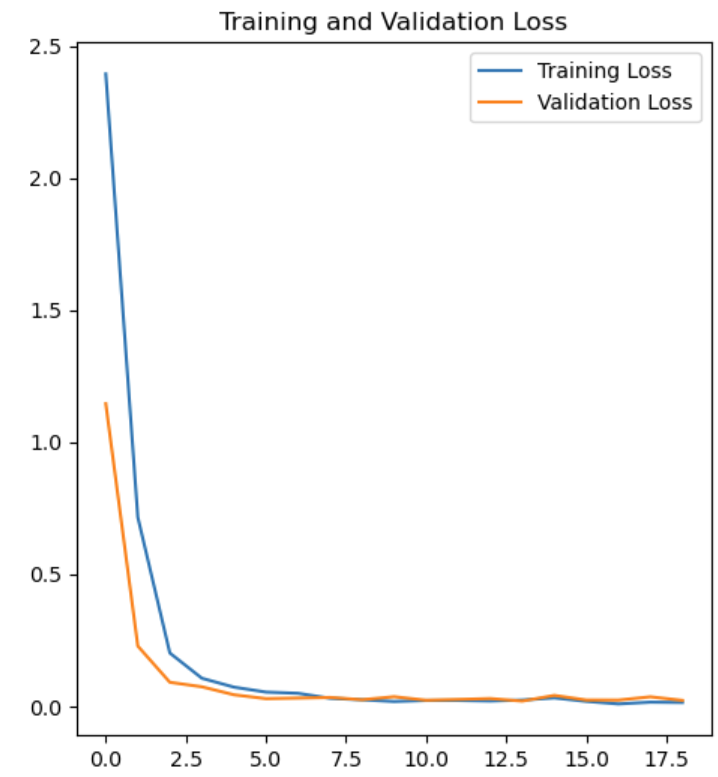}
    \caption{Training and Validation Loss VGG16- Deep Network}
    \label{fig:vgg16_results1}
\end{figure}

It is to be noted that the epochs mentioned are only till 18, as the early stopping mechanism of our system kicked in.
\vspace{5cm}
\subsection{VGG16 - Shallow Network}
The VGG16 shallow network we trained reduced the complexity of VGG16 Deep network by focusing the learining on the newly added layers rather than the base VGG16 layers.
Based, on the graphs below for training and validation accuracies as well as losses, it can be seen that the performance of the models was quite steady and consistent with what was expected. 
\begin{figure}[h]
    \centering
    \includegraphics[width=0.4\textwidth]{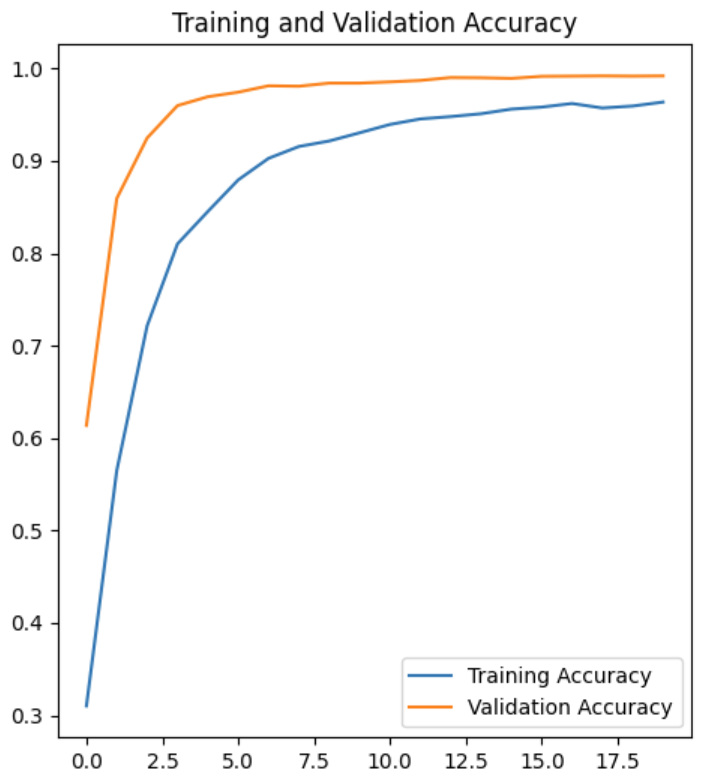}
    \caption{Training and Validation Accuracy VGG16- Shallow Network}
    \label{fig:vgg16_results2}
\end{figure}
\begin{figure}[h]
    \centering
    \includegraphics[width=0.4\textwidth]{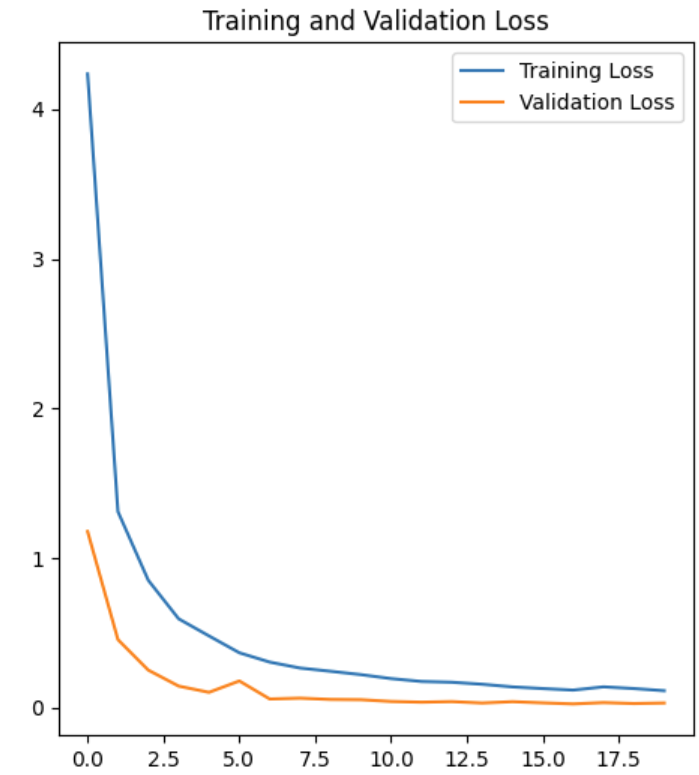}
    \caption{Training and Validation Loss VGG16- Shallow Network}
    \label{fig:vgg16_results3}
\end{figure}
It is to be noted that the epochs mentioned are only till 18, as the early stopping mechanism of our system kicked in.

\subsection{VGG16 - Fine Tuned Network - With and Without Batching}
This model froze the early layers while the top layers of the base VGG16 model are fine-tuned to balance the learning while ensuring not to alter the initial feature detection.

Both the batched as well as non-batched models perform well, and also while the non-batched version of the model does perform well, in terms of compute time, it take more training time to generate the same results as that of the batched version.

\begin{figure}[h]
    \centering
    \includegraphics[width=0.3\textwidth]{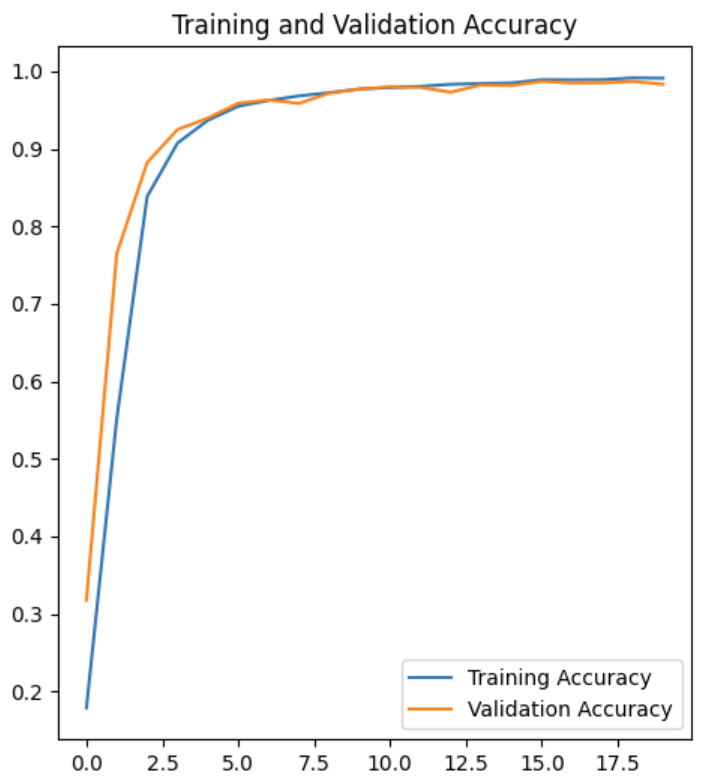}
    \caption{Training and Validation Accuracy VGG16- Fine Tuned- Batched}
    \label{fig:vgg16_results4}
\end{figure}
\begin{figure}[h]
    \centering
    \includegraphics[width=0.3\textwidth]{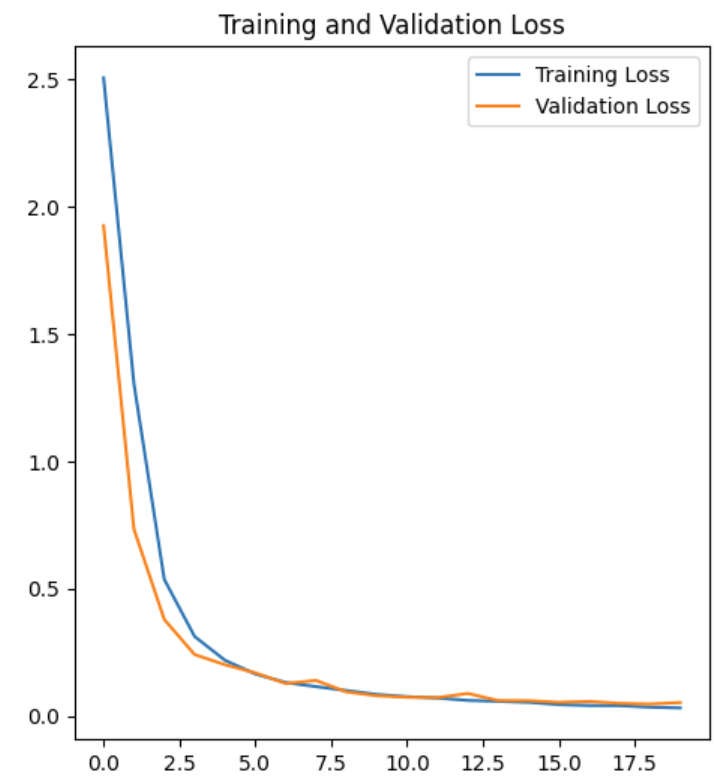}
    \caption{Training and Validation Loss VGG16- Fine Tuned- Batched}
    \label{fig:vgg16_results5}
\end{figure}
 
\begin{figure}[h]
    \centering
    \includegraphics[width=0.3\textwidth]{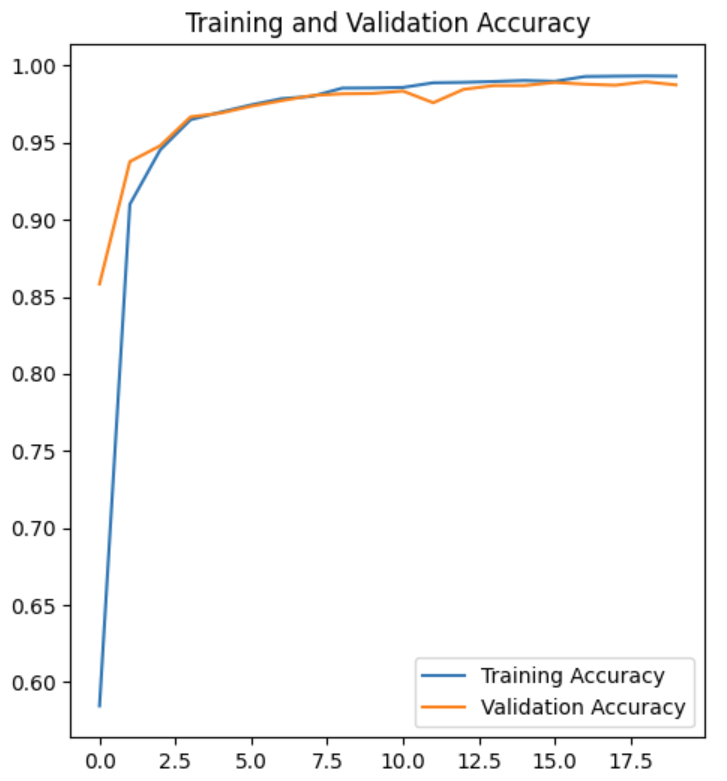}
    \caption{Training and Validation Accuracy VGG16- Fine Tuned- Non Batched}
    \label{fig:vgg16_results6}
\end{figure}
\begin{figure}[h]
    \centering
    \includegraphics[width=0.3\textwidth]{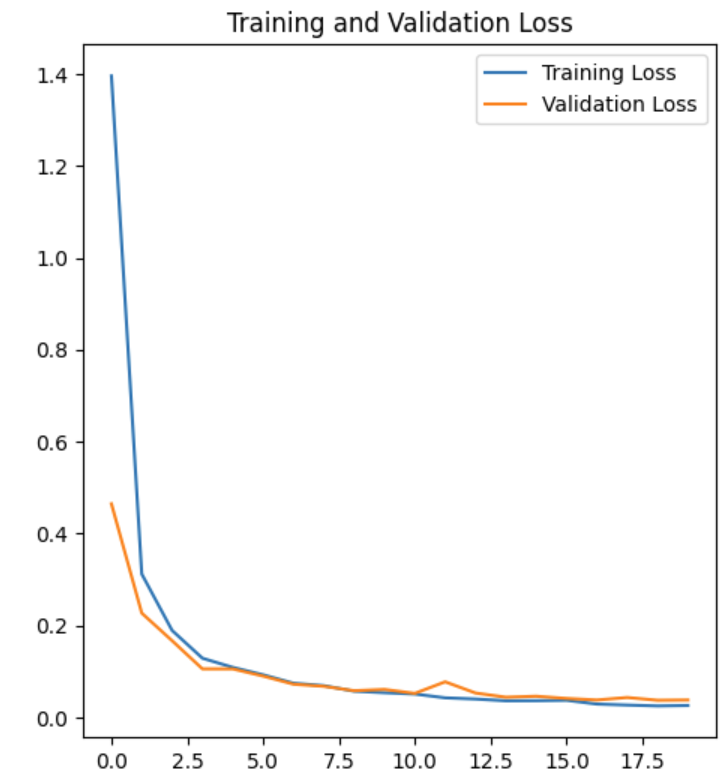}
    \caption{Training and Validation Loss VGG16- Fine Tuned- Non Batched}
    \label{fig:vgg16_results7}
\end{figure}

It is to be noted that the epochs mentioned are only till 18, as the early stopping mechanism of our system kicked in.
\vspace{10cm}
\subsection{VGG19- Deep Network}
The VGG19 models Deep Networks makes use of the base VGG19 model with additional flatten and dense layers in order to leverage a deeper network for particular traininig over the learning aspects of this dataset.

Based, on the graphs below for training and validation accuracies as well as losses, it can be seen that the performance of the models was quite steady and consistent with what was expected, however, it is worth mentioning that the training time for the same was quite drastically increased.

\begin{figure}[h]
    \centering
    \includegraphics[width=0.4\textwidth]{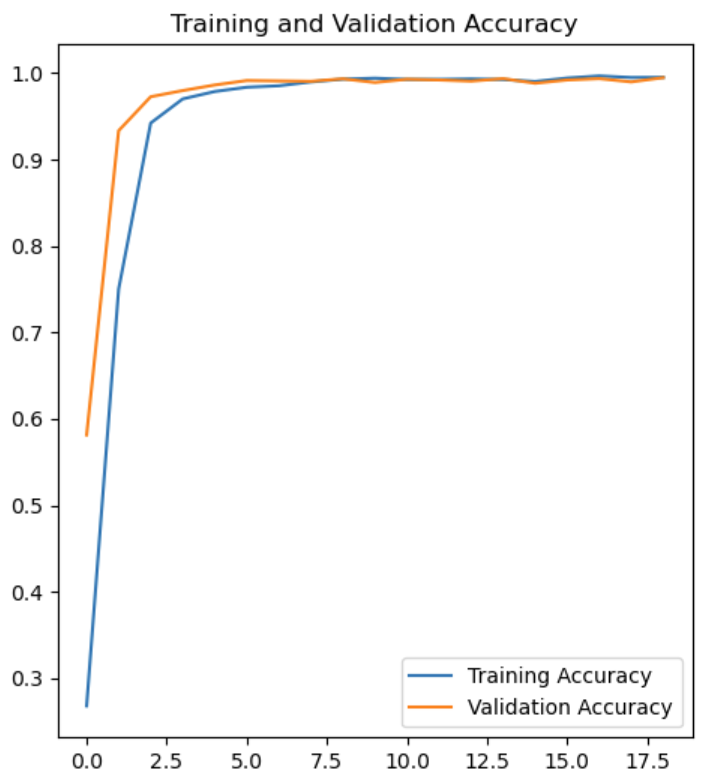}
    \caption{Training and Validation Accuracy VGG19- Deep Network}
    \label{fig:vgg16_results8}
\end{figure}
\begin{figure}[h]
    \centering
    \includegraphics[width=0.4\textwidth]{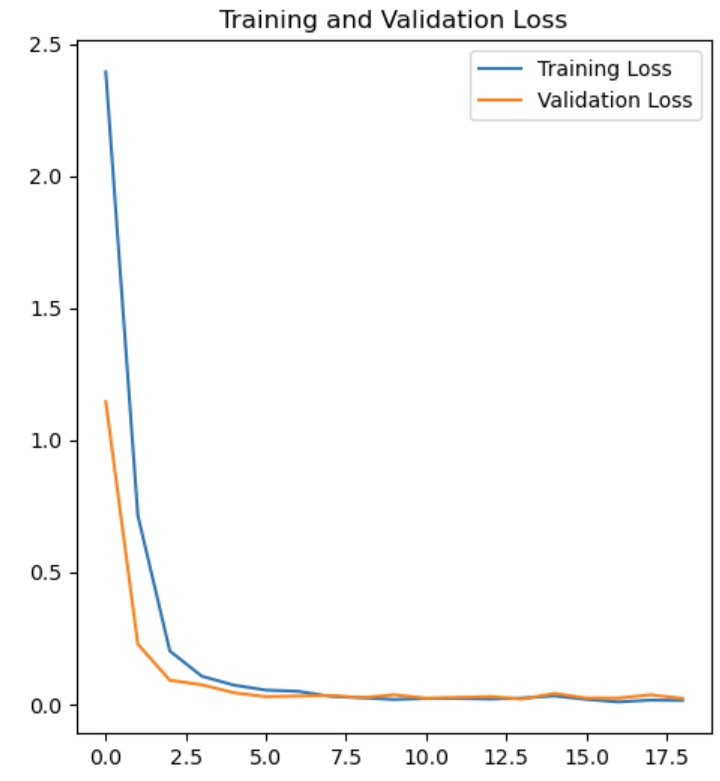}
    \caption{Training and Validation Loss VGG19- Deep Network}
    \label{fig:vgg16_results9}
\end{figure}

It is to be noted that the epochs mentioned are only till 18, as the early stopping mechanism of our system kicked in.

\subsection{VGG19 - Shallow Network}
The VGG19 shallow network we trained reduced the complexity of VGG19 Deep network by focusing the learining on the newly added layers rather than the base VGG19 layers.
Based, on the graphs below for training and validation accuracies as well as losses, it can be seen that the performance of the models was quite steady and consistent with what was expected. 
\begin{figure}[h]
    \centering
    \includegraphics[width=0.4\textwidth]{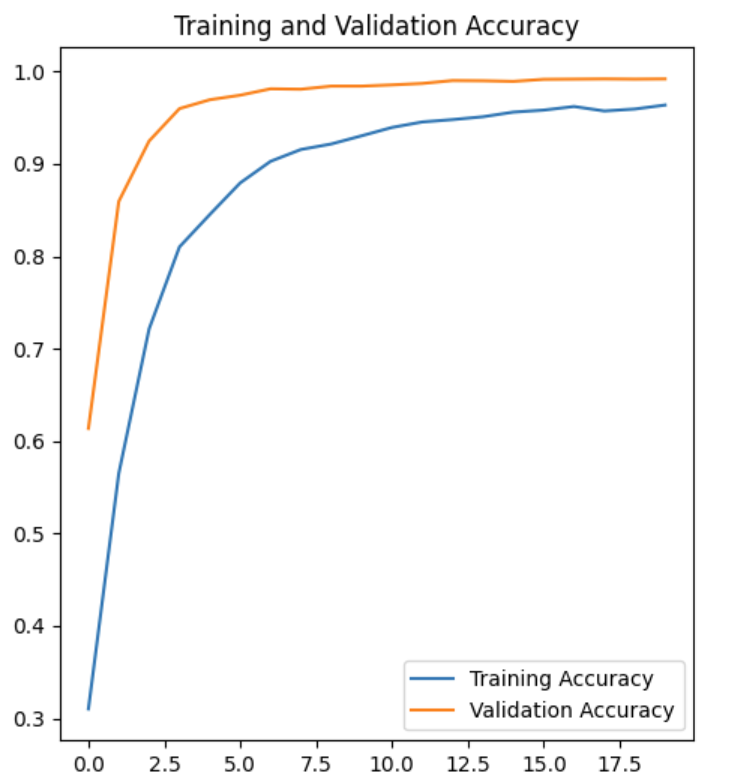}
    \caption{Training and Validation Accuracy VGG19- Shallow Network}
    \label{fig:vgg19_results}
\end{figure}
\begin{figure}[h]
    \centering
    \includegraphics[width=0.4\textwidth]{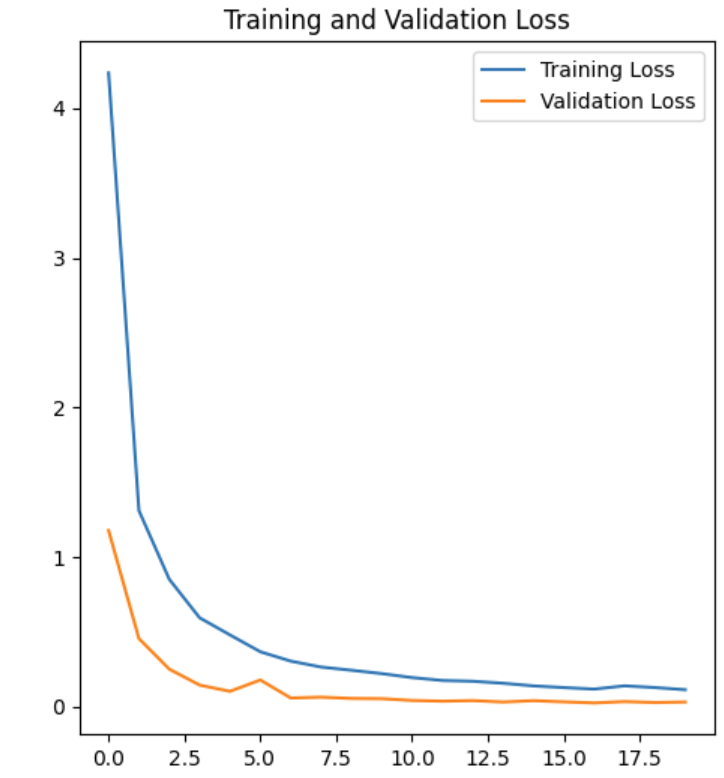}
    \caption{Training and Validation Loss VGG19- Shallow Network}
    \label{fig:vgg16_results10}
\end{figure}
It is to be noted that the epochs mentioned are only till 18, as the early stopping mechanism of our system kicked in.

\subsection{VGG16 - Fine Tuned Network - With and Without Batching}

This model froze the early layers while the top layers of the base VGG19 model are fine-tuned to balance the learning while ensuring not to alter the initial feature detection.

Both the batched as well as non-batched models perform well, and also while the non-batched version of the model does perform well, in terms of compute time, it take more training time to generate the same results as that of the batched version.

\begin{figure}[h]
    \centering
    \includegraphics[width=0.4\textwidth]{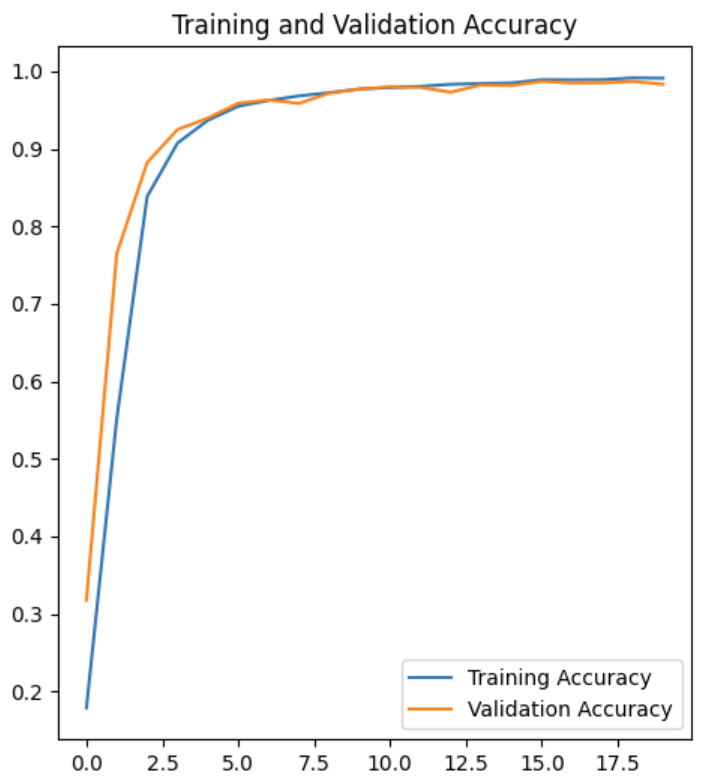}
    \caption{Training and Validation Accuracy VGG19- Fine Tuned- Batched}
    \label{fig:vgg16_results11}
\end{figure}
\begin{figure}[h]
    \centering
    \includegraphics[width=0.4\textwidth]{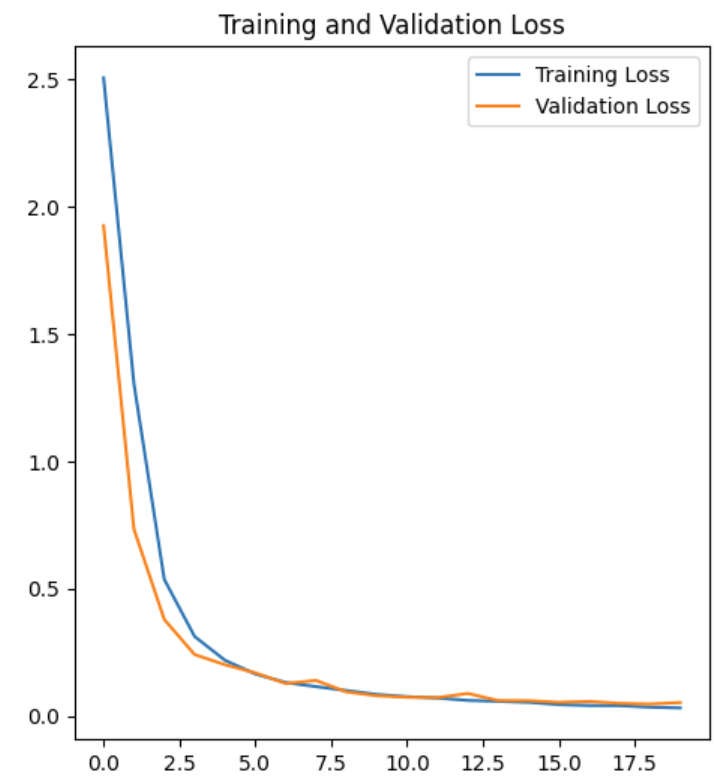}
    \caption{Training and Validation Loss VGG19- Fine Tuned- Batched}
    \label{fig:vgg16_results12}
\end{figure}
 
\begin{figure}[h]
    \centering
    \includegraphics[width=0.4\textwidth]{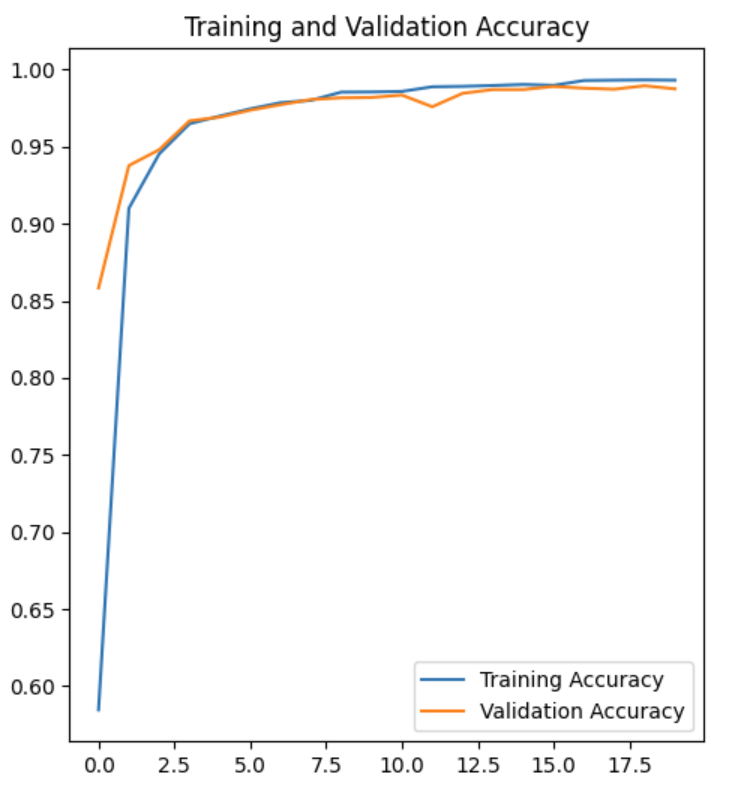}
    \caption{Training and Validation Accuracy VGG19- Fine Tuned- Non Batched}
    \label{fig:vgg16_results13}
\end{figure}
\begin{figure}[h]
    \centering
    \includegraphics[width=0.4\textwidth]{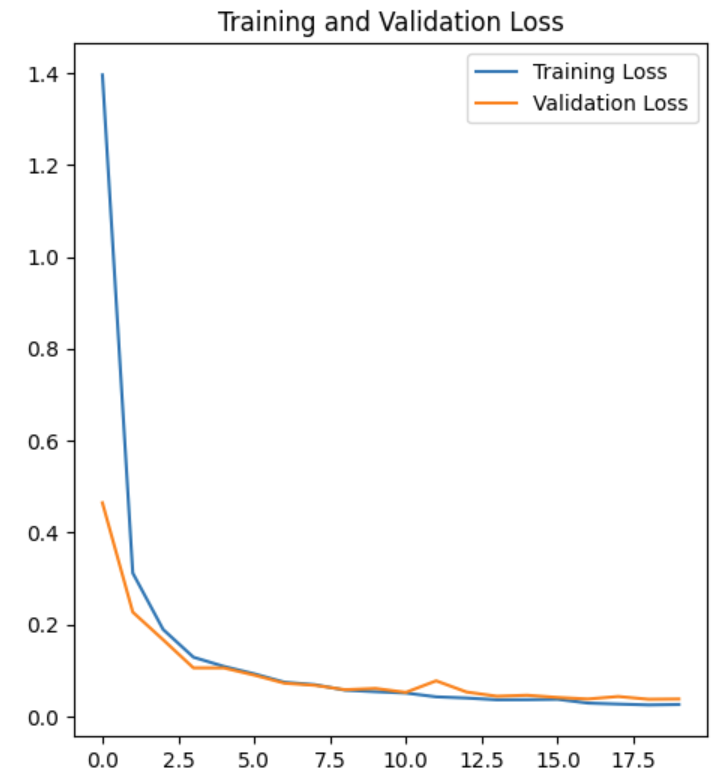}
    \caption{Training and Validation Loss VGG19- Fine Tuned- Non Batched}
    \label{fig:vgg16_results14}
\end{figure}

\vspace{7cm}

It is to be noted that the epochs mentioned are only till 18, as the early stopping mechanism of our system kicked in.
\vspace{5cm}
\subsection{Hybrid CNN-Transformer}
The hybrid architecture combine the best of the current two worlds when it comes to technology, utilizing a CNN backbone on a pre-trained ResNet50 along with a Transformer Encoder. The idea behind this is to test how computationally expensive can this be and can it  be a sustainable model.

The graphs below indicate quite interesting results, showcasing that even though we had set the initial number of epochs to 20, only 5 were required for the early stopping to kick in due to the performance of the model. Even though it has a shaky start on the training daqta in the first epoch, there was a sharp increase on accuracy and decrease in losses on the second epoch.

\begin{figure}[h]
    \centering
    \includegraphics[width=0.4\textwidth]{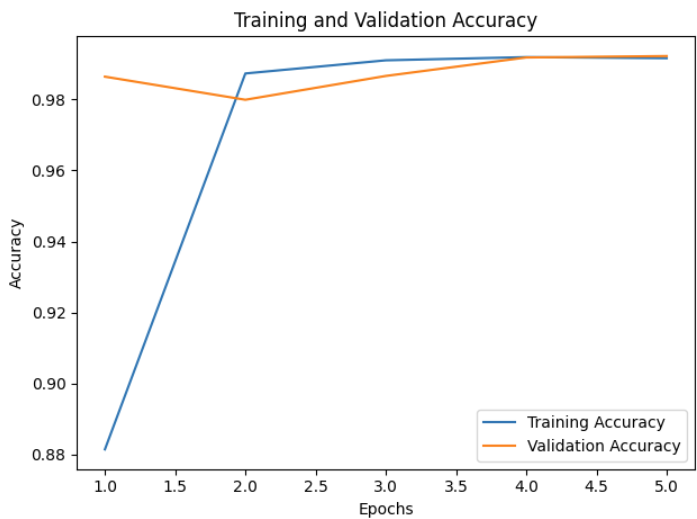}
    \caption{Training and Validation Accuracy for Hybrid CNN-Transformer}
    \label{fig:hybrid_cnn_transformer_results1}
\end{figure}
\begin{figure}[h]
    \centering
    \includegraphics[width=0.4\textwidth]{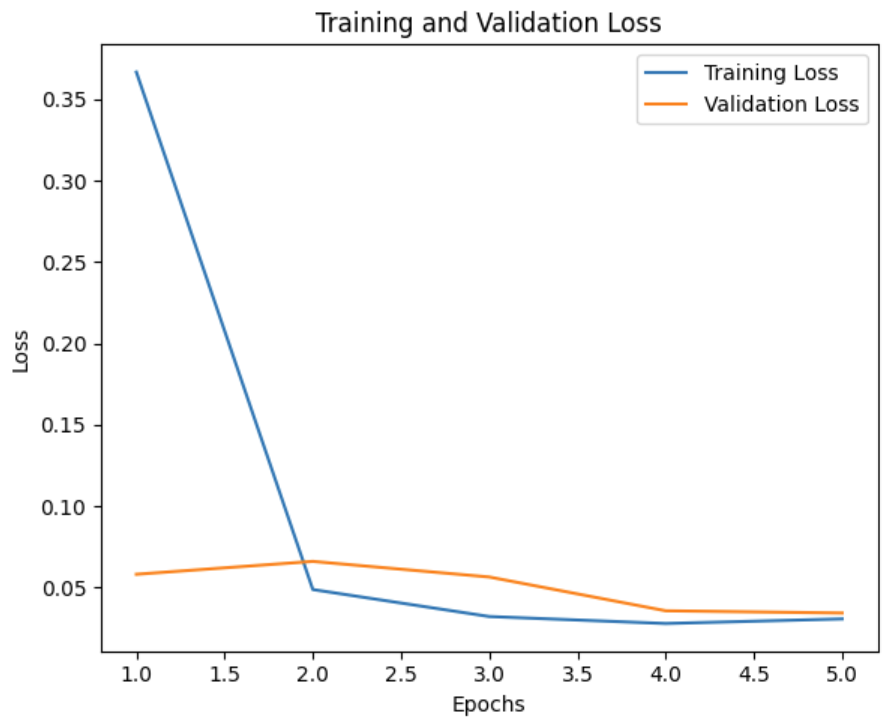}
    \caption{Training and Validation Loss for Hybrid CNN-Transformer}
    \label{fig:hybrid_cnn_transformer_results2}
\end{figure}

It is to be noted that the epochs mentioned are only till 5, as the early stopping mechanism of our system kicked in. This can be seen as quite an interesting result given that this hybrid mechanism took the least training to be able to achieve such results.

\section{Testing Results}

After the initial evaluation during training and validation, we subjected the models to a separate test dataset (derived from the same StateFarm Dataset) to evaluate their real-world applicability and robustness. This testing phase is crucial to verify the generalization capabilities of each model outside the controlled conditions of the training environment.

This test dataset was formulated by taking 10 random images from each of the 10 classes provided in the dataset in order to test two things:
\begin{itemize}
    \item \textbf{Accuracy of the Saved Models}
    \item \textbf{Time taken by models to evaluate the 100 test images}
\end{itemize}
The saved models were loaded one-by-one and evaluated on the same dataset.

The results can be seen below in tabular format:

\begin{table}[h!]
  \centering
  \caption{Model Performance: Accuracies and Elapsed Times}
  \label{tab:model_performance2}
  \begin{tabular}{lcc}
    \toprule
    Model & Accuracy & Elapsed Time (seconds) \\
    \midrule
    Simple CNN & 0.89 & 10.56 \\
    VGG16 Deep & 0.95 & 10.95 \\
    VGG16 Shallow & 0.93 & 9.14 \\
    VGG16 FT Batched & 0.96 & 9.07 \\
    VGG16 FT Non-Batched & 0.96 & 10.05 \\
    VGG19 Deep & 0.94 & 10.68 \\
    VGG19 Shallow & 0.94 & 9.68 \\
    VGG19 FT Batched & 0.98 & 8.89 \\
    VGG19 FT Non-Batched & 0.97 & 9.46 \\

    Hybrid CNN Transformer & 0.98 & 11.05 \\
    \bottomrule
  \end{tabular}
\end{table}
FT above is an abbrevation of Fine-Tuned

In order to showcase the above table as a graph, it would be better to visualize it as the same for better understanding:

\begin{figure}[h]
    \centering
    \includegraphics[width=0.4\textwidth]{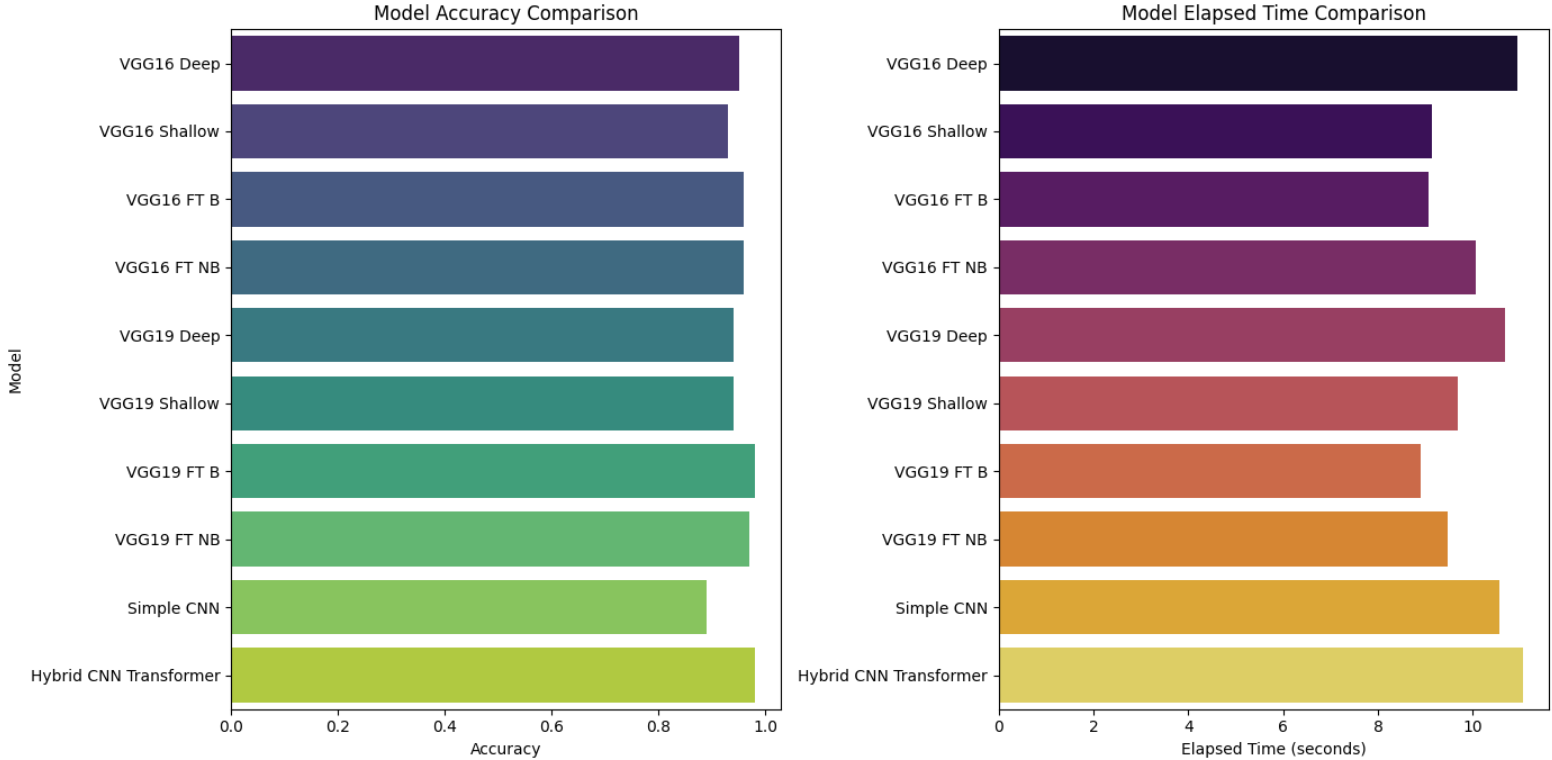}
    \caption{Overall Performance of Models}
    \label{fig:hybrid_cnn_transformer_results3}
\end{figure}

While the previous graph is quite insightful by visualizing the accuracy and elapsed time, it would be highly beneficial to visualize the two metrics in a single graph for better understanding.

\begin{figure}[h]
    \centering
    \includegraphics[width=0.4\textwidth]{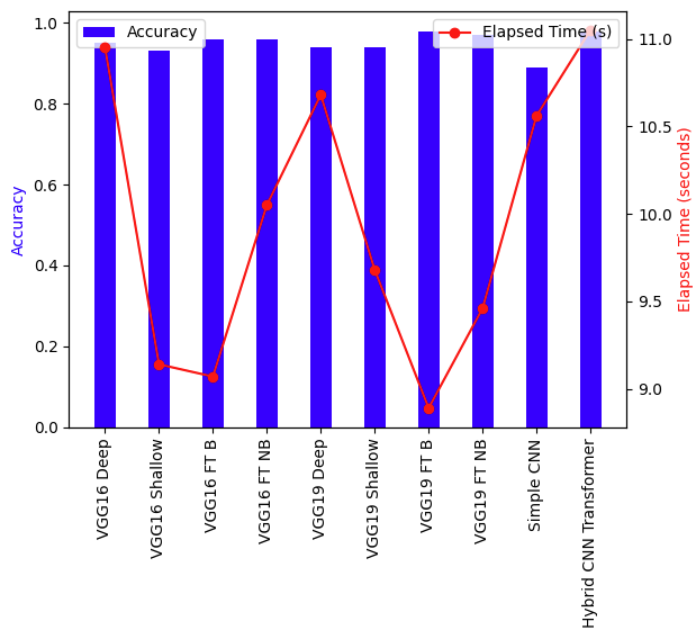}
    \caption{Accuracy and Speed of Models}
    \label{fig:hybrid_cnn_transformer_results4}
\end{figure}

From the above figures, it can be seen that the evaluation of various deep learning models against the Simple CNN baseline, it's clear that advanced models like VGG16, VGG19, and the Hybrid CNN Transformer not only improve in accuracy but also show distinct differences in processing times. The VGG models, particularly when fine-tuned, strike an effective balance between high accuracy and reasonable inference speeds. Notably, the fine-tuned VGG19 demonstrates the best optimization, achieving the highest accuracy with a comparatively low elapsed time. The Hybrid CNN Transformer, while slightly slower, matches this high accuracy. \textbf{These insights underscore the trade-offs between speed and accuracy in model selection, emphasizing the need for careful consideration based on specific application requirements.}

\section{Conclusion}
Our study examined several deep learning architectures to identify the most effective model for real-time driver distraction detection. The VGG19 FineTuned Batched model demonstrated the highest accuracy (0.98) among the tested models, closely followed by the Hybrid CNN Transformer, which also showed excellent performance (0.98 accuracy) but with slightly higher processing time. The results indicate a significant trade-off between model accuracy and evaluation time, which is critical for real-time applications where both accuracy and speed are crucial.

\subsection{Accuracy and Evaluation Time Trade-off}
Our findings showcase the importance of considering both accuracy and processing speed when selecting a model for deployment in real-world scenarios. While more complex models like the VGG19 FineTuned Batched and the Hybrid CNN Transformer offer high accuracy, their longer evaluation times may not be suitable for all real-time applications. On the other hand, simpler models such as the VGG16 Shallow provide faster evaluation times but at the cost of reduced accuracy.

\subsection{Hardware Capabilities}
The deployment of these models in a real-time environment also demands careful consideration of the hardware capabilities. Advanced models require powerful processing hardware, which may increase the cost and energy consumption of the deployed system. Therefore, balancing the hardware efficiency with model complexity is essential for practical implementations.
\vspace{0.5cm}

In conclusion, while the current models show promising results, there still remain a substantial scope for innovation in optimizing them for practical, real-time applications where both accuracy and speed are very important.

\end{document}